# Causal Discovery from a Mixture of Experimental and Observational Data

Gregory F. Cooper
Center for Biomedical Informatics
University of Pittsburgh
Pittsburgh, PA 15213
gfc@cbmi.upmc.edu

Changwon Yoo
Center for Biomedical Informatics
University of Pittsburgh
Pittsburgh, PA 15213
cwyoo@cbmi.upmc.edu

## Abstract

This paper describes a Bayesian method for combining an arbitrary mixture of observational and experimental data in order to learn causal Bayesian networks. Observational data are passively observed. Experimental data, such as that produced by randomized controlled trials, result from the experimenter manipulating one or more variables (typically randomly) and observing the states of other variables. The paper presents a Bayesian method for learning the causal structure and parameters of the underlying causal process that is generating the data, given that (1) the data contains a mixture of observational and experimental case records, and (2) the causal process is modeled as a causal Bayesian network. This learning method was applied using as input various mixtures of experimental and observational data that were generated from the ALARM causal Bayesian network. In these experiments, the absolute and relative quantities of experimental and observational data were varied systematically. For each of these training datasets, the learning method was applied to predict the causal structure and to estimate the causal parameters that exist among randomly selected pairs of nodes in ALARM that are not confounded. The paper reports how these structure predictions and parameter estimates compare with the true causal structures and parameters as given by the ALARM network.

## 1 INTRODUCTION

Causal knowledge makes up much of what we know and want to know in science. Thus, causal modeling and discovery are central to science and many other areas of inquiry.

Experimental studies, such as randomized controlled trials (RCTs), often provide the most trustworthy methods we have for establishing causal relationships from data. In an experimental study, one or more variables is manipulated (typically randomly) and the effects on other variables are measured. Such studies, while potentially highly informative, may not be safe, ethical, logistically feasible, or financially worthwhile.

Observational data is passively observed. Such data are more readily available than experimental data, and indeed, most databases are observational. As observational computer databases become increasingly available, opportunities increase for using them for causal discovery.

In general, there can exist both observational and experimental data on a set of variables of interest. For example, in clinical medicine there is a growing abundance of observational data contained in routinely collected electronic medical records. In addition, for selected variables of high clinical interest, there are data from RCTs. We need a coherent way of combining these two types of data to arrive at an overall assessment of the causal relationships among clinical variables.

Bayesian discovery of causal networks is an active field of research in which numerous advances have been—and are continuing to be—made in areas that include causal representation, model assessment and scoring, and model search (Cooper 1999).

In prior work on Bayesian discovery of causal networks, researchers have focused primarily on methods for discovering causal relationships from observational data. A notable exception is a paper by Heckerman on learning influence diagrams as causal models, which contains the essential ideas for learning causal Bayesian networks from a combination of experimental data under deterministic manipulation and observational data (Heckerman 1995). The contribution of the current paper is to use those ideas to investigate explicitly and in detail the learning of causal structures and parameters from an arbitrary mixture of observational and experimental data. In particular, the current paper presents a general Bayesian analysis of this learning task, including the situation in which experimental manipulation is not deterministic. The paper also specializes the general formulation to arrive at a closed-form Bayesian scoring metric for learning causal structures and parameters from a data mixture. Significantly, this scoring metric is a simple variation on a previous scoring metric for Bayesian network learning (Cooper and Herskovits 1992, Heckerman, et al. 1995). Thus, previous implementations of that metric can be readily adapted to learn causal networks from a combination of observational and experimental data. Finally, the paper investigates the learning performance of this metric when given a mixture of experimental and observational data that were generated from the ALARM causal Bayesian network.

## 2 MODELING METHOD

A causal Bayesian network (or *causal network* for short) is a Bayesian network in which each arc is interpreted as a direct causal influence between a parent node (variable) and a child node, relative to the other nodes in the network (Pearl 1988). Figure 1 illustrates the structure of a hypothetical causal Bayesian network structure, which contains five nodes. Due to limited space, the probabilities that are associated with this causal network structure are not shown.

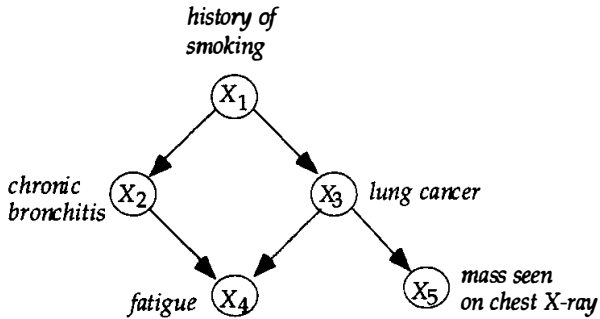

Figure 1: A hypothetical causal Bayesian network structure.

The causal network structure in Figure 1 indicates, for example, that a *history of smoking* can causally influence whether *lung cancer* is present, which in turn can causally influence whether a patient experiences *fatigue*.

The causal Markov condition gives the conditional independence relationships that are specified by a causal Bayesian network:

> A node is independent of its non-descendants (i.e., non-effects) given its parents (i.e., its direct causes).

The causal Markov condition permits the joint distribution of the $n$ variables in a causal Bayesian network to be factored as follows (Pearl 1988):

$$P(x_1, x_2, \ldots, x_n \mid K) = \prod_{i=1}^{n} P(x_i \mid \pi_i, K),$$

where $x_i$ denotes a state of variable $X_i$, $\pi_i$ denotes a joint state of the parents of $X_i$, and $K$ denotes background knowledge that is discussed in the next section.

### 2.1 A BAYESIAN ANALYSIS

This section considers the posterior probability that variable[1] $X$ causes variable $Y$ given database $D$ on measured variables $V$. Let $H$ denote an additional set of hidden (latent) variables. We use $V^+$ to designate the union of $V$ and $H$. Let $S$ denote an arbitrary causal Bayesian network structure containing all of the variables in $V^+$. Let $K$ denote our background knowledge that may influence our beliefs about the causal relationships among the variables in $V^+$. Such background knowledge could come from scientific laws, common sense, expert opinion, accumulated personal experience, as well as other sources. As we will see later in this section, $K$ also can contain knowledge of which cases in $D$ are experimental and which are observational. We can derive the posterior probability that $X$ causes $Y$ as

$$P(X \to Y \mid D, K) = \sum_{S:\, \{X \to Y\} \in S} P(S \mid D, K), \quad (1)$$

where the sum is taken over all causal network structures that (1) contain just the nodes in $V^+$, (2) contain an arc from $X$ to $Y$, and (3) have a non-zero prior probability.[2] Based on the properties of probabilities, the term within the sum in Equation 1 may be rewritten as follows:

$$P(S \mid D, K) = \frac{P(S, D \mid K)}{P(D \mid K)}$$

$$= \frac{P(S, D \mid K)}{\sum_S P(S, D \mid K)}. \quad (2)$$

Since relative to the entire set of causal structures being considered, the probability $P(D \mid K)$ is a constant, Equation 2 shows that the posterior probability of causal structure $S$ is proportional to $P(S, D \mid K)$, which we can view as a *score* of $S$ in the context of $D$. The probability terms on the right side of Equation 2 may be expanded as follows:

$$P(S, D \mid K) =$$
$$P(S \mid K) P(D \mid S, K) = \quad (3)$$
$$P(S \mid K) \int P(D \mid S, \theta_S, K) P(\theta_S \mid S, K) d\theta_S,$$

where (1) $P(S \mid K)$ is a prior belief that $S$ captures correctly the qualitative causal relationships among the variables in $V^+$, (2) $\theta_S$ are the probabilities (parameters) that relate the nodes in $S$ quantitatively to their respective parents, (3) $P(D \mid S, \theta_S, K)$ is the likelihood of data $D$ being produced given that the causal process generating the data is a causal Bayesian network given by $S$ and $\theta_S$, and (4) $P(\theta_S \mid S, K)$ expresses a belief about the probability distributions that serve to model the underlying causal process. The integral in Equation 3 integrates out the parameters $\theta_S$ in a causal Bayesian network with structure $S$ to derive $P(D \mid S, K)$, which is called the *marginal likelihood*. Combining Equations 1, 2, and 3, we obtain Equation 4.

---

[1] We use the terms *variable* and *node* interchangeably in this paper.
[2] In Equation 1 and subsequent equations in this section, the substructure $X \to Y$ can be replaced by any other substructure containing a subset of the variables in $V^+$. For simplicity of exposition, however, we focus in this section on a pairwise causal relationship of the form $X \to Y$.

$$P(X \rightarrow Y \mid D, K) = \frac{\sum_{S:\,\{X \rightarrow Y\} \in S} P(S \mid K) \int P(D \mid S, \theta_S, K) P(\theta_S \mid S, K) d\theta_S}{\sum_{S} P(S \mid K) \int P(D \mid S, \theta_S, K) P(\theta_S \mid S, K) d\theta_S} \quad (4)$$

The only assumption made in Equation 4 is the following:

*Assumption 1.* Causal relationships are represented using causal Bayesian networks.

The full Bayesian approach to causal discovery expressed by Equation 4 considers—at least in principle—all causal Bayesian networks that are *a priori* possible. Thus, for example, the sums in Equation 4 are over all possible causal Bayesian network structures on $V^+$, and the integrals are over all possible parameters for each possible causal structure. The result of such a global analysis of causality is that the derived posterior probabilities summarize a comprehensive, normative belief about the causal relationships among a set of variables.

The Bayesian analysis given by Equation 4 presents three considerable challenges: (1) the assessment of prior probabilities on causal network structures and parameters, (2) the summation over a large set of causal network structures, and (3) the evaluation of the integral. In this paper, we focus on task 3. In particular, we introduce a set of assumptions that simplify the evaluation of the integral, and we show that the solution corresponds closely to a previous solution for observational data only.

*Assumption 2.* The cases in $D$ are a random sample from the joint probability distribution given by a causal Bayesian network $B$ with structure $S$ and parameters $\theta_S$.

Assumption 2 implies that cases are independent, *conditioned on the generating model*, which is specified by Assumption 1. Using Assumptions 1 and 2, we can express the integral in Equation 3 as follows:

$$P(D \mid S, K) = \int [\prod_{h=1}^{m} P(C_h \mid S, \theta_S, K)] \, P(\theta_S \mid S, K) d\theta_S, \quad (5)$$

where each $C_h$ represents one of $m$ cases in dataset $D$. In Equation 5, the term $P(C_h \mid \theta_S, S, K)$ denotes using the causal Bayesian network with structure $S$ and parameters $\theta_S$ to infer the probability of the state of the variables as given by $C_h$. If $C_h$ corresponds to a case containing only observational data, then inference is performed using the entire Bayesian network. If the generation of case $C_h$ involved the experimental manipulation of one or more model variables, then we make the following additional assumption:

*Assumption 3.* For each experimentally manipulated variable $X_i$ in case $C_h$, the probability $P(C_h \mid \theta_S, S, K)$ is modeled by removing from $B$ the arcs into $X_i$, and setting $P(X_i = k \mid K) = 1$, where $k$ is the value to which $X_i$ was manipulated.

The justification for Assumption 3 is as follows. If $X_i$ is being manipulated deterministically by forces outside the causal network model, then $X_i$ is no longer under the influence of any variables in the model, and thus, the arcs into $X_i$ should be removed. $P(X_i = k \mid K)$ is set equal to 1, because we assume that experimental manipulation of $X_i$ is deterministic. See (Spirtes, et al. 1993, Section 3.7.2) for a detailed discussion of Assumption 3.

Given Assumption 3, consider a case $C_h$ that contains a variable $X_i$ that is manipulated to state $k$.[3] For this case, the term $P(C_h \mid \theta_S, S, K)$ in Equation 5 is inferred as follows: modify $S$ by removing the arcs into $X_i$, remove the parameters in $\theta_S$ that correspond to the removed arcs in $S$, set $P(X_i = k \mid K) = 1$, then use this modified causal Bayesian network to infer the probability of the state of the variables in $C_h - \{X_i\}$ (all of which are observational variables). The generalization to the simultaneous manipulation of multiple variables is straightforward.

When there are hidden variables or missing data, the Bayesian approach can model them explicitly and normatively (Cooper 1993); however, exact computation of the integral in Equation 5 with current methods is usually intractable, even when a causal network contains only one hidden variable. The use of sampling and approximation methods have shown promise in estimating the integral when there is missing data or hidden variables. Modeling hidden variables and missing data, however, is not a focus of the current paper. Therefore, we will only consider the case in which all model variables are set to known states:

*Assumption 4.* There are no missing data or hidden variables.

Given Assumption 4, Bayesian network inference reduces to computing a product of conditional probabilities. Therefore, Equation 5 simplifies to be:

---

[3] Formally, we assume the information that $X_i$ is manipulated to state $k$ in case $h$ is contained in the background knowledge denoted by $K$.

$$P(D \mid S, K) =$$

$$\int [\prod_{h=1}^{m} \prod_{i=1}^{n} P(x_i^h \mid \pi_i^h, \theta_S, K)] \, P(\theta_S \mid S, K) d\theta_S \quad (6)$$

where there are $n$ variables in $V$, $x_i^h$ denotes the state of variable $X_i$ in case $h$, and $\pi_i^h$ denotes the states of the parents of $X_i$ in case $h$. When $h$ is a case in which $X_i$ is manipulated, then the term $P(x_i^h \mid \pi_i^h, \theta_S, K)$ in Equation 6 is replaced by 1, since according to Assumption 3 it is with probability 1 that variable $X_i$ will be in the state to which it is manipulated. If $X_i$ is not manipulated, but rather observed, then $P(x_i^h \mid \pi_i^h, \theta_S, K)$ is the probability given by $\theta_S$ for the state of $X_i$ conditioned on the states of its parents in case $h$. Notice that some of the parents of observational variable $X_i$ may themselves have been experimentally manipulated to the states they have in case $h$. Note also that $X_i$ could have been manipulated in some cases in $D$ and not manipulated in other cases.

In order to efficiently evaluate the integral in Equation 6, researchers have introduced several assumptions that lead to a closed form solution (Cooper and Herskovits 1992, Heckerman, et al. 1995). Under the assumptions that follow in this section, as expressed in (Geiger and Heckerman 1995), the integral in Equation 6 can be computed efficiently in closed form.

*Assumption 5.* Variables are discrete.

*Assumption 6* (parameter independence)
*Global parameter independence*: For each causal Bayesian network structure, the parameters (probabilities) associated with one node are probabilistically independent of the parameters associated with other nodes.
*Local parameter independence*: The parameters associated within a node given one instance of its parents are independent of the parameters of that node given other instances of its parent nodes.[4]

Assumptions 5 and 6 permit the terms in Equation 6 to be regrouped to obtain the following equation:

$$P(D \mid S, K) = \quad (7)$$

$$\int [\prod_{i=1}^{n} \prod_{j=1}^{q_i} \prod_{k=1}^{r_i} P(x_i = k \mid \pi_i = j, \theta_S, K)^{N_{ijk}}] P(\theta_S \mid S, K) d\theta_S$$

where $r_i$ is the number of states that $X_i$ can have, $q_i$ denotes the number of joint states that the parents of $X_i$ can have, and $N_{ijk}$ is the number of cases in $D$ in which node $X_i$ is <u>passively observed</u> to have state $k$ when its parents have states as given by $j$. Thus, for example, if $X_i$ were manipulated in all $m$ cases in $D$, then $N_{ijk}$ would equal 0 for all states of $j$ and $k$. In addition, we note that the tally of $N_{ijk}$ is indifferent to how the *parents* of $X_i$ attained their states as given by $j$.

We now introduce two additional assumptions.

*Assumption 7.* (parameter modularity (Heckerman, et al. 1995)) If a node has the same parents in two distinct networks, then the distribution of the parameters associated with this node are identical in both networks.

*Assumption 8.* The prior distribution of parameters associated with each node is Dirichlet.

Assumptions 7 and 8 permit replacing $P(\theta_S \mid S, K)$ in Equation 7 with a Dirichlet prior distribution. The solution to Equation 7 under Assumptions 7 and 8 is as follows (Cooper and Herskovits 1992, Heckerman, et al. 1995):

$$P(D \mid S, K) =$$

$$\prod_{i=1}^{n} \prod_{j=1}^{q_i} \frac{\Gamma(\alpha_{ij})}{\Gamma(\alpha_{ij} + N_{ij})} \prod_{k=1}^{r_i} \frac{\Gamma(\alpha_{ijk} + N_{ijk})}{\Gamma(\alpha_{ijk})} \quad (8)$$

where $\Gamma$ is the gamma function, $\alpha_{ijk}$ and $\alpha_{ij}$ express parameters of the Dirichlet prior distributions, and $N_{ij} = \sum_{k=1}^{r_i} N_{ijk}$.

Given Assumptions 1-8, it also follows from the results in (Cooper and Herskovits 1992, Heckerman, et al. 1995) that when $X_i$ is observed, we estimate its conditional distribution as follows:

$$P(x_i = k \mid \pi_i = j, D, S, K) = \frac{\alpha_{ijk} + N_{ijk}}{\alpha_{ij} + N_{ij}}. \quad (9)$$

Since the probabilities given by Equation 9 define all the parameters for a causal Bayesian network $B$ with structure $S$, we can use $B$ to perform probabilistic inference. Let $P(X_a \mid x_b, D, S, K)$ designate a generic instance of using $B$ to infer the distribution of the variables in $X_a$ conditioned on the state of other variables as given by $x_b$. We assume that $X_a$ contains only variables that will be observed. The variable set $x_b$ may, however, contain both observed and manipulated variables. For each manipulated variable in $x_b$, Assumption 3 implies that we remove the arcs into $x_b$ before inferring the conditional distribution of $X_a$.

We can infer the distribution of $X_a$ conditioned on $x_b$ by model averaging over all possible network structures as follows:

$$P(X_a \mid x_b, D, K) =$$

$$\sum_S P(X_a \mid x_b, D, S, K) P(S \mid D, K), \quad (10)$$

---

[4] Heckerman uses the term *mechanism independence* for the causal version of global parameter independence, and the term *component independence* for the causal version of local parameter independence (Heckerman 1995).

where the term $P(S \mid D, K)$ is obtained by substituting Equation 8 into Equation 3 and then substituting Equation 3 into Equation 2.

## 2.2 AN EXAMPLE

In this section we derive the marginal likelihood $P(D \mid S, K)$ for the causal structure $S$ equal to $X_1 \to X_2$, when given the data in Table 1, which contains 11 cases. There are two binary variables, $X_1$ and $X_2$. Each case has a state for each variable. If a state was obtained by observation, it appears in normal font (T or F). If a state was obtained by manipulation it appears as an outlined font ($\mathbb{T}$ or $\mathbb{F}$).

Table 1: An example dataset.

| $X_1$ | $X_2$ |
|---|---|
| T | T |
| T | F |
| T | T |
| F | F |
| F | T |
| $\mathbb{T}$ | T |
| $\mathbb{F}$ | F |
| T | $\mathbb{T}$ |
| F | $\mathbb{T}$ |
| T | $\mathbb{F}$ |
| F | $\mathbb{F}$ |

For this example, with $S$ equal to the causal structure $X_1 \to X_2$, we apply Equation 8 to derive $P(D \mid S, K)$ as follows:

$P(D \mid S, K) =$

$$\prod_{i=1}^{n} \prod_{j=1}^{q_i} \frac{\Gamma(\alpha_{ij})}{\Gamma(\alpha_{ij} + N_{ij})} \prod_{k=1}^{r_i} \frac{\Gamma(\alpha_{ijk} + N_{ijk})}{\Gamma(\alpha_{ijk})} =$$

$$\frac{\Gamma(\alpha_{11})}{\Gamma(\alpha_{11} + N_{11})} \frac{\Gamma(\alpha_{111} + N_{111})}{\Gamma(\alpha_{111})} \frac{\Gamma(\alpha_{112} + N_{112})}{\Gamma(\alpha_{112})} \cdot$$

$$\frac{\Gamma(\alpha_{21})}{\Gamma(\alpha_{21} + N_{21})} \frac{\Gamma(\alpha_{211} + N_{211})}{\Gamma(\alpha_{211})} \frac{\Gamma(\alpha_{212} + N_{212})}{\Gamma(\alpha_{212})} \cdot$$

$$\frac{\Gamma(\alpha_{22})}{\Gamma(\alpha_{22} + N_{22})} \frac{\Gamma(\alpha_{221} + N_{221})}{\Gamma(\alpha_{221})} \frac{\Gamma(\alpha_{222} + N_{222})}{\Gamma(\alpha_{222})} =$$

$$\frac{\Gamma(1)}{\Gamma(1+9)} \frac{\Gamma(1/2+4)}{\Gamma(1/2)} \frac{\Gamma(1/2+5)}{\Gamma(1/2)} \cdot$$

$$\frac{\Gamma(1/2)}{\Gamma(1/2+3)} \frac{\Gamma(1/4+2)}{\Gamma(1/4)} \frac{\Gamma(1/4+1)}{\Gamma(1/4)} \cdot$$

$$\frac{\Gamma(1/2)}{\Gamma(1/2+4)} \frac{\Gamma(1/4+1)}{\Gamma(1/4)} \frac{\Gamma(1/4+3)}{\Gamma(1/4)} = 5.97 \times 10^{-7},$$

where we have assumed that the parameters $a_{1jk}$ equal 1/2 and the parameters $a_{2jk}$ equal 1/4.[5] We use the convention that F is represented by $k = 1$ and T by $k = 2$. When $i = 2$, the parent state $X_1 = $ F of $X_2$ is represented by $j = 1$, and the parent state $X_1 = $ T of $X_2$ by $j = 2$. For example, consider the term $N_{222}$, which corresponds to the frequency with which jointly $X_2$ has the state T and $X_1$ (the parent of $X_2$ in the hypothesis being considered) has the state T. Following the analysis in Section 2.1, we derive $N_{222}$ from Table 1 by considering only the cases in which $X_2$ was *not* manipulated; these correspond to the first seven cases in the table. Of those seven cases, three occur when jointly $X_2$ has the state T and $X_1$ has the state T. Thus, $N_{222} = 3$ in the equation of the example.

Finally, note that if we assume that $P(X_1 \to X_2 \mid K) = 1/3$, we can solve Equation 3 as $P(S, D \mid K) = 1/3 \times 5.97 \times 10^{-7} = 1.99 \times 10^{-7}$, which can be used in solving Equation 2. □

## 2.3 NONDETERMINISTIC MANIPULATION

Section 2.1 considers the situation in which an experimental manipulation is deterministic. That is, when the experimenter decides to manipulate $X$ to state $x$, variable $X$ assuredly takes on state $x$. In the current section, we generalize this result to the situation in which manipulation may not be deterministic. A classic example in medicine is that a patient, who has volunteered to participate in a study, may be randomized to receive medication $d$, but he or she may decide not to take $d$.

Let $M_i$ be a variable that represents the value $k$ (from 1 to $r_i$) to which the experimenter wishes to manipulate $X_i$. Let $M_i = o$ denote that the experimenter does not wish to manipulate $X_i$, but rather, wants merely to observe its value. Augment the model variables $V$ in Section 2.1 to include $M_i$. Finally, carry out the analysis in Section 2.1 assuming only observational data. The causal network hypotheses used in that analysis will include probabilities that specify prior beliefs about the causal influence of $M_i$ on $X_i$. Those prior beliefs will be updated by data on stated experimental wishes and observed variable outcomes.

The general formulation described in the previous paragraph simplifies to the model in Section 2.1 when we assume that (1) with probability 1 variable $M_i$ is a parent of $X_i$, and (2) if $M_i = k$ in case $h$, then the term $P(x_i^h \mid \pi_i^h, \theta_S, K)$ in Equation 6 is replaced by $P(X_i^h = k \mid (\pi_i^h, M_i^h = k), \theta_S, K) = 1$, where, as before, $\pi_i^h$ represents the state of the hypothesized parents (among the variables in V) of $X_i$ in case h. In the formulation in the previous paragraph, in general the distribution $P(x_i^h \mid (\pi_i^h, M_i^h = k), \theta_S, K)$ is not deterministic. Moreover, prior belief and/or data may in some instances even support strongly that $M_i$ is not a parent of $X_i$, indicating that experimental intentions have little or no effect on the actual values of $X_i$.

---

[5] Section 3 contains an explanation of these choices for the $\alpha_{ijk}$ parameters.

## 3 EXPERIMENTAL METHODS

In evaluating causal learning, we ideally would know the real-world causal relationships (both the structure and parameters) among a set of variables of interest. With such knowledge we could generate experimental and observational data. Using these datasets as input, a learning method could predict the causal structure and estimate the causal parameters that exist among the modeled variables. These predictions and estimates would then be compared to the true causal relationships. Since confident knowledge of underlying causal processes is relatively rare, in this initial study of causal discovery from mixed data we used as a gold standard a causal model that was constructed by an expert. In particular, we used the ALARM causal Bayesian network[6], which contains 46 arcs and 37 nodes that can have from two to four possible states. Beinlich constructed ALARM as a research prototype to model potential anesthesia problems in the operating room (Beinlich, et al. 1989). In constructing ALARM, he used knowledge from the medical literature, as well as personal experience as an anesthesiologist. The remainder of this section describes how we generated data from ALARM and then used this data in evaluating the learning method described in Section 2.1.

### 3.1 DATA GENERATION

The 37 nodes in ALARM may be paired in 666 unique ways. We will denote an arbitrary pair of nodes as (X, Y). If there is at least one directed causal path from X to Y or from Y to X, we say that X and Y are *causally related*. If X and Y share a common ancestor, we say that X and Y are *confounded*. Table 2 summarizes the types of causal relationships among the 666 node pairs of ALARM.

We randomly selected 100 of the 666 node pairs. Table 3 shows the frequencies of the types of pairs that were sampled. The frequencies in Table 3 closely match those in Table 2, supporting that this sample of 100 is not biased.

In the experiment reported here, we focused only on node pairs that were not confounded. We did so to simplify the experimental design and analysis in this initial experiment. Table 4 shows the three possible modeled relationships that can exist between two nodes that are not confounded: (1) there is one or more causal paths from X to Y (H1), (2) there is one or more causal paths from Y to X (H2), or (3) X and Y have no d-connecting paths (Pearl 1988) between each other (H3).

---

[6] In particular, we used the version of ALARM that is publicly available for downloading as alarm.dsc from the Bayesian Network Repository at http://www-nt.cs.berkeley.edu/home/nir/public_html/Repository/alarm.htm.

Table 2: Types of node pairs in ALARM.

|  |  | confounded | | total |
|---|---|---|---|---|
|  |  | yes | no |  |
| causally related | Yes | 56 (8.4%) | 167 (25.1%) | 223 (33.5%) |
|  | No | 78 (11.7%) | 365 (54.8%) | 443 (66.5%) |
| total |  | 134 (20.1%) | 532 (79.9%) | 666 (100%) |

Table 3: Types of node pairs sampled from ALARM.

|  |  | confounded | | total |
|---|---|---|---|---|
|  |  | yes | no |  |
| causally related | Yes | 8 (8.0%) | 29 (29.0%) | 37 (37.0%) |
|  | No | 11 (11.0%) | 52 (52.0%) | 63 (63.0%) |
| total |  | 19 (19.0%) | 81 (81.0%) | 100 (100%) |

Table 4: The three unconfounded causal hypotheses being modeled. The double-headed arcs convey that the causal influence is either direct (relative to the modeled variables) or indirect.

| Hypothesis label | Causal Bayesian network hypothesis |
|---|---|
| H1 | X ⟶⟶ Y |
| H2 | X ⟵⟵ Y |
| H3 | X    Y |

From Table 2 we see that unconfounded nodes make up 79.9 percent of the node pairs in ALARM. Table 3 indicates that 81 unconfounded pairs were included in the sample; let U denote these 81 pairs. For each pair (X, Y) in U, we used stochastic simulation (Henrion 1988) to generate three types of data from ALARM. In particular, we generated data in which (1) X is manipulated and Y is observed, (2) Y is manipulated and X is observed, and (3) $X$ and $Y$ are both observed. For data that were generated under manipulation, we used a uniform prior over the states to which to manipulate the manipulated variable (e.g., if $X$ is binary with states T and F, then $P(manip(X = T)) = P(manip(X = F)) = 0.5)$.

## 3.2 LEARNING METHOD

For learning, a dataset $D$ consisted of $m/2$ cases in which $X$ was manipulated and $Y$ was observed, $m/2$ cases in which $Y$ was manipulated and $X$ was observed, and $n$ cases in which $X$ and $Y$ were both observed. Thus, $D$ contains $m + n$ cases. We varied $m$ incrementally from 0 to 500. For each value of $m$, we varied $n$ incrementally from 0 to 500.

For each of the 81 pairs of unconfounded nodes, we used the method in Section 2.1 to compute a posterior probability distribution over the three causal network structures in Table 4. We assumed a uniform prior probability of 1/3 for each structure. In applying Equation 8 to derive the marginal likelihood, we used the following parameter priors: $a_{ijk} = 1/(q_i r_i)$ for all $i$, $j$ and $k$. This choice of priors has two properties (among others) (Heckerman, et al. 1995). First, the priors are weak in the sense that the marginal likelihood is influenced largely by the dataset $D$. Second, given only observational data, structures H1 and H2 will have equal posterior probabilities. These two properties provide a type of non-informative parameter prior. We chose to use a non-informative prior in the current evaluation in order to draw insights that are based mainly on the data and not on our own subject beliefs.

## 3.3 EVALUATION METRICS

For a given node pair $(X, Y)$, let $H_{true}$ designate which of the three structures from Table 4 is the relationship between $X$ and $Y$ in ALARM. For each pair $(X, Y)$ and dataset $D$, we derived the following structural error metric:

$$SErr_{X,Y}(D) = 1 - P(H_{true} | D, K),$$

where $P(H_{true} | D, K)$ is the posterior probability derived by using the method in Section 2. If that method always predicted the true relationship in ALARM with probability 1, then the error would be 0. We computed an overall structural error rate by averaging over all the pairwise error rates as follows:

$$SErr(D) = \frac{\sum_{X,Y} SErr_{X,Y}(D)}{b},$$

where the sum is taken over $b$ node pairs. In our analyses, $b$ is either 29 (corresponding to unconfounded, causally related nodes, as tabulated in Table 3) or 52 (corresponding to unconfounded, causally unrelated nodes).

We also were interested in how well the learned models are able to accurately predict the distribution of one variable given manipulation or observation of the other variable. In the remainder of this section, we define an error of predicting the distribution of $Y$ given that $X$ is observed. We also define an error of predicting the distribution of $Y$ given that $X$ is manipulated.

Let $x$ denote an arbitrary state of $X$ and $y$ an arbitrary state of $Y$. Let $P_A(X = x)$ denote the marginal probability that $X$ is observed to be $x$, according to the ALARM Bayesian network. Let $P_A(Y = y | X = x)$ designate a conditional probability as inferred using ALARM of observing $Y$ to be $y$ given that $X$ is observed to be $x$. Let $P_E(Y = y | X = x)$ be an estimate of the same conditional probability that is obtained by applying model averaging using Equation 10. For a given $(X, Y)$ we define as follows the expected error of predicting the observation of $Y$ given an observation of $X$:

$$OPErr_{X,Y}(D) = \sum_x P_A(X = x) \cdot$$

$$[\frac{1}{r_Y} \sum_y |P_A(Y = y | X = x) - P_E(Y = y | X = x)|],$$

where the outer sum is taken over all the states of $X$, the inner sum is taken over all states of $Y$, and $r_Y$ is the number of states that $Y$ can have. For a given database $D$, $OPErr_{X,Y}(D)$ measures the expected absolute error of predicting an observed state of $Y$ given an observed state of $X$. The expectation is taken with respect to the observational distribution of the states of $X$. The overall observational prediction error ($OPErr$) is defined as follows:

$$OPErr(D) = \frac{\sum_{X,Y} OPErr_{X,Y}(D)}{b}.$$

Now consider the situation in which $X$ is *manipulated* to a state and we observe the distribution of $Y$. We use the notation $manip(X = x)$ to represent that $X$ is manipulated to the state $x$. We have no particular reason to assume that $X$ would be manipulated to any one state more than another. Thus, for the purpose of deriving an error metric, we assume that $X$ is equally likely to be manipulated to each of its $r_X$ possible states. Under these terms, we obtain the following manipulation error metrics:

$$MPErr_{X,Y}(D) = \sum_x \frac{1}{r_X} [\frac{1}{r_Y} \cdot$$

$$\sum_y |P_A(Y = y | manip(X = x)) - P_E(Y = y | manip(X = x))|].$$

$$MPErr(D) = \frac{\sum_{X,Y} MPErr_{X,Y}(D)}{b}.$$

## 4 EXPERIMENTAL RESULTS AND DISCUSSION

Table 5a shows the results of the structure prediction error when $X$ and $Y$ are causally related. Since the prior probability of each structure is 1/3, the error rate is 2/3, or approximately 0.667, when there are no data. On average, the more experimental data that $D$ contains, the more probable is the prediction of the generating causal relationship. Although the experiment stopped at 500 experimental cases, we would expect the error rate to continue decreasing as more experimental cases were available. Observational data alone are not sufficient to determine whether $X$ is causing $Y$ or $Y$ is causing $X$. Significantly, however, the table indicates that observational data can augment experimental data in decreasing the error (i.e., increasing the posterior probability assigned to the "true" data-generating relationship). Observational data is able to do so by helping "eliminate" hypothesis H3 that $X$ and $Y$ are not related.

Table 5a: The structural error metric $SErr(D)$ for pairs of nodes that are causally related but not confounded (H1 and H2) for different combinations of observational data and data resulting from experimental manipulation.

Experimental data ($m$ cases)

| Obs. data ($n$ cases) | 0 | 50 | 100 | 300 | 500 |
|---|---|---|---|---|---|
| 0 | 0.667 | 0.508 | 0.502 | 0.349 | 0.300 |
| 50 | 0.701 | 0.482 | 0.429 | 0.314 | 0.278 |
| 100 | 0.716 | 0.491 | 0.428 | 0.306 | 0.268 |
| 300 | 0.693 | 0.439 | 0.391 | 0.281 | 0.230 |
| 500 | 0.693 | 0.419 | 0.365 | 0.273 | 0.209 |

As stated above, when there is no experimental or observational data, the error rate in Table 5a is 0.667 (= 2/3). When observational data is added, the error rate initially increases, because 11 of the 29 generating structures are only very weakly correlated; with samples on the order of 50 to 500 cases, these 11 structures are given relatively low posterior probabilities of being causally related. When we performed additional simulation with up to 50,000 cases of observational data, in 10 of the 11 pairs the error rate converged to 0.5 as expected.

Table 5b shows the structure prediction error when $X$ and $Y$ are not causally related. As expected, both experimental and observational data are able to determine about equally well that $X$ and $Y$ are independent. Combining the two types of data decreases the error, as anticipated.

Table 5b: The structural error metric $SErr(D)$ for pairs of nodes that are not confounded and not related (H3).

Experimental data ($m$ cases)

| Obs. data ($n$ cases) | 0 | 50 | 100 | 300 | 500 |
|---|---|---|---|---|---|
| 0 | 0.667 | 0.333 | 0.257 | 0.103 | 0.089 |
| 50 | 0.291 | 0.242 | 0.220 | 0.087 | 0.081 |
| 100 | 0.235 | 0.193 | 0.185 | 0.088 | 0.077 |
| 300 | 0.119 | 0.111 | 0.112 | 0.077 | 0.072 |
| 500 | 0.106 | 0.101 | 0.102 | 0.074 | 0.071 |

In Tables 6a and 6b, a given number of observational cases alone yields a lower observational prediction-error rate than the same number of experimental cases alone. A primary reason for this result appears to be that all the observational data are relevant in performing parameter estimation for observational prediction, whereas in general only a subset of experimental data is relevant. Notice also that with 500 observational cases, adding experimental data is relatively ineffective in further lowering the error rate.

Table 6a: The observational prediction-error metric $OPErr(D)$ for pairs of nodes that are causally related and not confounded (H1 and H2).

Experimental data ($m$ cases)

| Obs. data ($n$ cases) | 0 | 50 | 100 | 300 | 500 |
|---|---|---|---|---|---|
| 0 | 0.301 | 0.055 | 0.045 | 0.029 | 0.020 |
| 50 | 0.042 | 0.036 | 0.031 | 0.025 | 0.018 |
| 100 | 0.031 | 0.028 | 0.025 | 0.019 | 0.017 |
| 300 | 0.019 | 0.017 | 0.017 | 0.014 | 0.013 |
| 500 | 0.016 | 0.015 | 0.015 | 0.013 | 0.012 |

Table 6b: The observational prediction-error metric $OPErr(D)$ for pairs of nodes that are not confounded and not related (H3).

Experimental data ($m$ cases)

| Obs. data ($n$ cases) | 0 | 50 | 100 | 300 | 500 |
|---|---|---|---|---|---|
| 0 | 0.311 | 0.057 | 0.037 | 0.018 | 0.015 |
| 50 | 0.037 | 0.029 | 0.024 | 0.016 | 0.014 |
| 100 | 0.022 | 0.020 | 0.019 | 0.013 | 0.012 |
| 300 | 0.014 | 0.012 | 0.013 | 0.010 | 0.009 |
| 500 | 0.011 | 0.010 | 0.010 | 0.009 | 0.008 |

Table 7a indicates that when $X$ and $Y$ are causally related, the sole use of experimental data leads to lower errors for predicting manipulations than does the sole use of observational data. Unlike experimental data, observational data cannot distinguish whether $X$ is causing $Y$ or $Y$ is causing $X$. Table 7a shows that when there are small amounts of experimental data, observational data can significantly decrease the prediction error. For example, with 50 experimental cases and no observational data, the error is 0.056. Adding just 100 observational cases decreases the error by about a half to 0.029. With 50 experimental cases and 300 observational cases, the error reduces to 0.015, which is comparable to the error of 0.019 when using 500 experimental cases alone. This pattern is interesting, because in the real world we often are stuck with a relatively small amount of experimental data (because it is expensive and difficult to acquire), but we may have an abundance of observational data.

It is noteworthy that in Table 7a the error rate is only 0.047 when using 500 observational cases alone to make experimental predictions. We believe this result occurred (at least in part) because many of the causal relationships being analyzed are weak. This issue, however, deserves additional investigation.

Table 7a: The manipulation prediction-error metric *MPErr(D)* for pairs of nodes that are causally related but not confounded (H1 and H2).

| Obs. data (*n* cases) | Experimental data (*m* cases) | | | | |
|---|---|---|---|---|---|
| | 0 | 50 | 100 | 300 | 500 |
| 0 | 0.285 | 0.056 | 0.042 | 0.026 | 0.019 |
| 50 | 0.060 | 0.038 | 0.032 | 0.025 | 0.018 |
| 100 | 0.056 | 0.029 | 0.025 | 0.020 | 0.017 |
| 300 | 0.048 | 0.015 | 0.015 | 0.014 | 0.013 |
| 500 | 0.047 | 0.013 | 0.012 | 0.012 | 0.011 |

Table 7b shows that when $X$ and $Y$ are not causally related, then observational and experimental data are similar in terms of predicting the effect of manipulation (namely, no effect), particularly for the larger datasets.

Table 7b: The manipulation prediction-error metric *MPErr(D)* for pairs of nodes that are not confounded and not related (H3).

| Obs. data (*n* cases) | Experimental data (*m* cases) | | | | |
|---|---|---|---|---|---|
| | 0 | 50 | 100 | 300 | 500 |
| 0 | 0.311 | 0.050 | 0.036 | 0.018 | 0.015 |
| 50 | 0.036 | 0.029 | 0.024 | 0.016 | 0.014 |
| 100 | 0.021 | 0.020 | 0.018 | 0.013 | 0.012 |
| 300 | 0.013 | 0.012 | 0.013 | 0.010 | 0.009 |
| 500 | 0.011 | 0.010 | 0.010 | 0.009 | 0.008 |

## 5 CONCLUSIONS AND FUTURE WORK

A Bayesian analysis of causal discovery from a mixture of experimental and observational data is similar to an analysis for observational data alone. Under assumptions, a closed-form Bayesian scoring metric was derived that differs from an existing scoring metric (for observational data alone) only in the interpretation of the numerical counts (i.e., the $N_{ijk}$ terms). This means that existing implementations of that observational scoring metric can be readily adapted to score causal networks when both experimental and observational data are available.

An empirical evaluation of the learning method was performed using data generated from the ALARM causal Bayesian network. General patterns that were observed support that combining observational and experimental data is useful when learning causal structure and when performing predictions based on observations and based on manipulations. For those tasks, the results quantify how data of one type tends to be most useful (in lowering error rates) when data of the other type is relatively scarce.

We emphasize, however, that these patterns have been investigated and seen thus far only for data generated from the ALARM network. Additional studies are needed to see if these patterns also appear when using data generated from other causal networks.

Future work beyond the current paper includes expanding the set of hypotheses to model confounding of two variables, which would increase the number of causal hypotheses. We have begun investigating causal learning when confounding is possible; our results, however, are still preliminary. In addition, the evaluation reported in the current paper uses simple pairwise causal relationships. Observational data potentially can be much more informative when using three or more measured variables. We plan to use ALARM to investigate causal discovery when causal hypotheses are allowed to contain more than two variables.

It will be important to evaluate causal learning using a variety of different causal Bayesian networks as generating models. It also will be interesting to apply these learning methods with real observational and experimental data. Using real data will likely require that we be able to model missing data, some of which may not be missing at random.

### Acknowledgments

We thank Stefano Monti for his comments on an earlier draft of this paper. The research reported here was supported by NSF grant IIS-9812021.

Causal Discovery from a Mixture of Data    125## References

Beinlich, I.A., Suermondt, H.J., Chavez, R.M. and Cooper, G.F. (1989) The ALARM monitoring system: A case study with two probabilistic inference techniques for belief networks, In: *Proceedings of the Second European Conference on Artificial Intelligence in Medicine* 247–256.

Cooper, G.F. (1993) A method for learning belief networks that contain hidden variables, In: *Proceedings of the Workshop on Knowledge Discovery in Databases* 112–124.

Cooper, G.F. (1999) An overview of the representation and discovery of causal relationships using Bayesian networks. In: Glymour C. and Cooper G.F. (Eds.), *Computation, Causation, and Discovery* (AAAI Press and MIT Press, Menlo Park, CA).

Cooper, G.F. and Herskovits, E. (1992) A Bayesian method for the induction of probabilistic networks from data, *Machine Learning* 9 309-347.

Geiger, D. and Heckerman, D. (1995) A characterization of the Dirichlet distribution with application to learning Bayesian networks, In: *Proceedings of the Conference on Uncertainty in Artificial Intelligence* 196-207.

Heckerman, D. (1995) A Bayesian approach to learning causal networks, In: *Proceedings of the Conference on Uncertainty in Artificial Intelligence* 285-295.

Heckerman, D., Geiger, D. and Chickering, D. (1995) Learning Bayesian networks: The combination of knowledge and statistical data, *Machine Learning* 20 197-243.

Henrion, M. (1988) Propagating uncertainty in Bayesian networks by logic sampling. In: Lemmer J.F. and Kanal L.N. (Eds.), *Uncertainty in Artificial Intelligence 2* (North-Holland, Amsterdam) 149–163.

Pearl, J. (1988) *Probabilistic Reasoning in Intelligent Systems* (Morgan Kaufmann, San Mateo, CA).

Spirtes, P., Glymour, C. and Scheines, R. (1993) *Causation, Prediction, and Search* (Available at http://hss.cmu.edu/html/departments/philosophy/TETRAD.BOOK/book.html).